\documentclass[11pt]{article}

\usepackage[margin=1in]{geometry}
\usepackage{booktabs}
\usepackage{graphicx}
\usepackage{amsmath}
\usepackage{amssymb}
\usepackage{natbib}
\usepackage[hidelinks]{hyperref}
\usepackage{microtype}
\usepackage{array}
\usepackage{caption}
\usepackage{xcolor}

\usepackage{setspace}
\newcommand{\prtxt}{\mathrm{PR}}

\title{\textbf{When Do Attention Circuits Form?\\
Developmental Trajectories of Capability and Attention-Sink
Emergence Across Three 1B-Class Architectures}\thanks{Correspondence: \texttt{abbyxu@gmail.com}. Code, data, and reproducibility scripts: \url{https://github.com/skydancerosel/spectral-probe-circuits}}}

\author{Yongzhong Xu}

\date{}

\begin{document}

\maketitle

\begin{abstract}
\noindent
We track the developmental trajectory of attention-head circuit formation
across three 1B-class language models that span two architecture families
(dense transformer, mixture-of-experts) and two pretraining corpora
(The Pile, DCLM): Pythia 1B, OLMo 1B-0724-hf, and OLMoE 1B-7B-0924. At
each of 10 log-spaced revisions per model --- 30 mechanistic-interpretability
runs in total --- we apply a participation-ratio (PR) spectral signal and an
all-head capability-specific selectivity screen to identify induction,
previous-token, and BOS-attractor heads as they emerge.

Six findings. \textbf{(F1)} Layers 0 and 1 produce zero BOS-classified heads
at every revision in every model: the L0/L1 zero-BOS floor is an architectural
property, not a learned outcome. \textbf{(F2)} The whole-model BOS-attractor
fraction follows three distinct emergence shapes --- a gradual ramp in Pythia
1B (Pile, dense), a sharp phase transition in OLMo 1B (DCLM, dense, jumping
from 7\% to 70\% between adjacent checkpoints), and a gradual ramp in OLMoE
1B-7B (DCLM, MoE). \textbf{(F3)} In DCLM models, induction-circuit formation
precedes BOS-attractor formation by 10--20$\times$ in tokens (OLMo: induction
at 23B, BOS-50\% at 264B; OLMoE: induction at 20B, BOS-50\% at 419B).
Capability-circuit formation and attention-sink formation are two transitions,
not one. \textbf{(F4)} The capability-specific screen recovers two-thirds
of the final-checkpoint induction circuit within 0.3--2\% of total training
tokens (a recall measure); circuit identification does not require the final
model. \textbf{(F5)} For every
final-checkpoint induction head sampled across all three models, per-head PR
is elevated \emph{at or before} the first revision at which that head crosses
its capability-selectivity threshold. \textbf{(F6)} The induction circuit's
membership is not stable across training: screening every head independently
at each revision, the formation-time circuit overlaps the final circuit by
only $\sim$1/3 (Jaccard $0.29$--$0.33$), as strong early induction heads decay
out while others are recruited late --- F4's early \emph{recall} does not imply
early circuit \emph{identity}. Group ablation makes the turnover causal and
locates its sign in the \emph{fate} of each decaying head: heads that become
attention sinks are causally retired, whereas heads that become
duplicate-token heads stay load-bearing (duplicate-detection is the
prior-occurrence step of the induction composition), so the
selectivity screen under-identifies the causal circuit.

The results refine the ``induction phase transition'' framing
\citep{olsson2022}: in 1B-class models trained on DCLM, the induction
transition and the attention-sink transition are separated by an order of
magnitude in tokens, and they have qualitatively different emergence shapes.
The L0/L1 zero-BOS floor and the data $\times$ architecture interaction in
BOS-attractor dynamics are constraints any mechanistic theory of attention
sinks must explain. We position this paper as the activation-space half of
an emergence story whose parameter-space half is studied through
spectral-edge dynamics \citep{xu2026spectraledge}.
\end{abstract}

\section{Introduction}

\citet{olsson2022} identified induction heads --- attention heads
implementing the $A\,B\,\ldots\,A\to B$ copy pattern --- as a key
mechanistic feature of in-context learning in transformer language models,
and connected their formation to a sharp ``phase transition'' visible as a
bump in the loss curve during pretraining. \citet{xiao2024streaming} described a
distinct empirical phenomenon: pretrained LMs reliably allocate large
attention probability to the first (BOS) token regardless of content, a
behaviour they termed an \emph{attention sink}, which can be exploited for
KV-cache compression during streaming inference. Both phenomena are
described in terms of attention patterns; both are observed in modern
decoder-only language models; and both are sometimes folded together under
the heading of ``attention emergence during pretraining.''

This paper asks: when, exactly, do these two phenomena form during training
of 1B-class language models, and are they the \emph{same} transition? We
answer this empirically by tracking capability circuits (induction,
previous-token) and the BOS attractor side-by-side across three 1B-class
models that span two architecture families and two pretraining corpora.

\paragraph{Models.} Pythia 1B \citep{biderman2023}, a GPT-NeoX dense
transformer trained on The Pile \citep{gao2020pile}; OLMo 1B-0724-hf
\citep{groeneveld2024}, a Llama-style dense transformer trained on DCLM
\citep{li2024dclm}; and OLMoE 1B-7B-0924 \citep{muennighoff2024}, a
Llama-style mixture-of-experts transformer (1B active / 7B total
parameters, 64 experts, top-8 routing) trained on DCLM. These three models
factorize cleanly: Pythia vs OLMo isolates the corpus (Pile vs DCLM) at
fixed scale and dense architecture; OLMo vs OLMoE isolates the architecture
(dense vs MoE) at fixed corpus.

\paragraph{Method.} For each model we use the three-step recipe of the
companion methodology paper \citep{paper1_methodology}: per-head
participation ratio computed on a synthetic induction batch, an all-head
capability-specific selectivity screen, and group ablation. We apply the
screen at 10 log-spaced revisions per model --- 30 runs in total ---
producing a developmental panel.

\paragraph{Findings.} Six findings emerge, summarised in the abstract.
The central result is that in DCLM-trained 1B models, the induction
transition (Olsson's phenomenon) and the BOS-attractor transition
(Xiao's phenomenon) are separated by 10--20$\times$ in tokens and have
different emergence shapes (gradual ramp vs sharp jump). The two
phenomena cannot be identified with a single phase-transition event at
1B scale on DCLM data.

\paragraph{Position.} This paper is the \emph{activation-space} half of a
broader study of emergence in pretraining. The companion methodology paper
\citep{paper1_methodology} defines the recipe and validates it at the
final checkpoint; this paper applies the recipe developmentally; the
companion pattern-selectivity paper \citep{paper3_circuits} studies
cross-task generalisation. A complementary parameter-space program ---
spectral-edge dynamics \citep{xu2026spectraledge}, with a toy-scale
connection drawn between spectral-edge geometry and the Linear Centroids
Hypothesis \citep{walker2026lch} by \citet{xu2026gradsens} --- studies
the same emergence event in weight-space rather than activation-space. We
return to the bridge between these windows in the Discussion.

\section{Related Work}

\paragraph{Induction heads and the phase transition.}
\citet{olsson2022} introduced the induction-head abstraction and reported a
``phase transition'' in pretraining co-located with a bump in the loss
curve, a sudden growth in in-context-learning capability, and the
appearance of attention heads implementing the prefix-match-and-copy
pattern. Their analysis used integrated-gradients-style attribution on
small ($\le\!\!100\mathrm{M}$-parameter) decoder-only models. The induction
phase transition has since become a standard reference point for
emergence-during-pretraining claims. This paper revisits the
\emph{transition} part of that claim at 1B scale across data and
architecture.

\paragraph{Attention sinks.}
\citet{xiao2024streaming} documented attention sinks --- the observation that
trained decoder-only LMs reliably attend to the first token regardless of
content --- and used the phenomenon as a substrate for streaming inference
via fixed first-token KV retention. They did not study the formation
dynamics of sinks during pretraining. We do: BOS-attractor heads are
identified by the same selectivity screen as capability circuits, and
their per-revision counts give a token-resolved formation curve.

\paragraph{Pythia, OLMo, OLMoE.}
\citet{biderman2023} released Pythia precisely to enable cross-checkpoint
analyses of pretraining dynamics, with 154 checkpoints per model size
spanning random-init to 300B tokens. \citet{groeneveld2024} and
\citet{muennighoff2024} released OLMo and OLMoE with similar checkpoint
discipline. The 30-revision panel in this paper depends on all three
projects' checkpoint releases. We document checkpoint-availability
limitations (OLMo's earliest is 2B tokens; OLMoE's is 20B tokens) and
their effect on the resolved developmental curve in Section~\ref{sec:setup}.

\paragraph{Cross-architecture mechanistic transfer.}
Prior work has documented that the specific circuits a model uses for a
given task differ across architectures and scales
\citep{lieberum2023,wang2022ioi,marks2024sparse}. The cross-architecture
result in this paper is different: we hold the synthetic capability fixed
(induction on a controlled batch) and ask whether the \emph{timing} of
circuit formation, and the timing of the co-occurring BOS-attractor
phenomenon, transfer across pretraining corpora and architectures.

\paragraph{Automated circuit discovery.}
\citet{conmy2023} introduced ACDC, an iterative edge-pruning algorithm for
post-hoc circuit identification. The recipe used here is complementary and
predates a fully-trained model: a spectral signal that can be read at any
intermediate checkpoint, plus a per-class selectivity screen. We use ACDC
as motivation for the design rather than as a baseline.

\paragraph{Parameter-space emergence.}
A complementary parameter-space program studies the same emergence event
from a different vantage point. \citet{xu2026spectraledge} tracks the
singular-value (spectral-edge) geometry of weight-matrix trajectories.
\citet{xu2026gradsens} relates that geometry, at toy-model scale, to the
Linear Centroids Hypothesis \citep{walker2026lch}, via
gradient-direction-sensitive features whose coupling to spectral-edge
dynamics is otherwise hidden by optimizer trajectories. Our PR signal is
computed on activations, not parameters; the parameter-side work treats
the parameter spectrum of the same checkpoints. We discuss in
Section~\ref{sec:discussion} what an integrated account might look like.

\section{Methodology Recap}
\label{sec:method}

The recipe is defined in detail in \citet{paper1_methodology}. We summarise
here only what is needed to interpret the developmental results.

\paragraph{Synthetic induction batch.} 2000 sequences of length 256, RNG
seed 42. Each sequence has the structure $[\text{filler}]\,A\,B\,
[\text{filler}]\,A$, with $A$ and $B$ random tokens drawn from vocabulary
IDs $[100,\,10{,}000)$. The induction prediction is $B$ at the position
immediately following the second occurrence of $A$.

\paragraph{Per-head participation ratio.} For each (layer $L$, head $H$,
revision $t$), extract the per-head attention output at the second-$A$
position over the batch, giving $M \in \mathbb{R}^{N \times d_\text{head}}$
(here $N=2000$). Compute the singular spectrum $\{\sigma_i\}$ of $M$ and
the participation ratio
\[
  \prtxt(L,H,t) \;=\; \exp\!\big(H(p)\big),
  \quad p_i = \sigma_i^2 / \textstyle\sum_j \sigma_j^2,
\]
where $H(p)=-\sum_i p_i \log p_i$. The trajectory feature is the integral
$I(L,H) = \sum_t \max(\prtxt(L,H,t)-1,\,0)\cdot\Delta \log(\text{tokens}_t)$.
Throughout this paper we plot per-revision $\max(\prtxt(L,H,t)-1,0)$ as the
\emph{integrand} of $I$, so that temporal structure is preserved.

\paragraph{Capability-pattern screen.} For each head, compute attention
from the query position to canonical target positions for six classes:
\textsc{induction}, \textsc{previous-token}, \textsc{duplicate-token},
\textsc{first-token / bos}, \textsc{self}, and \textsc{local}. Selectivity
is (target attention)/(uniform-other baseline). A head is classified into
the class with maximum selectivity if that maximum exceeds 30$\times$
(class assignment); it is admitted to a class circuit if selectivity for
that class exceeds 50$\times$ (induction, BOS) or 100$\times$
(previous-token).

\paragraph{All-head capability-specific screen.} In the
attention-sink-dominated regime ($\ge\!70\%$ of heads classify as
\textsc{first-token} at the assignment threshold; see Section~\ref{sec:f2}),
ranking by best class surfaces BOS-attractor heads ahead of capability
heads. For circuit identification of a specific target capability $X$, we
instead take all heads with $\mathrm{sel}_X \ge \tau_X$ regardless of best
class. This is the screen used for the per-revision induction-circuit
identification in Sections~\ref{sec:f3}, \ref{sec:f4}, and~\ref{sec:f5}.

\paragraph{Causal verification.} Group-ablate the screen-identified circuit
by zeroing the per-head slice of the residual contribution at the attention
output projection. Matched-random controls share layers and head count but
not identities. Top-1 accuracy on the synthetic induction batch is the
primary metric. Final-checkpoint circuit ablations are documented in the
companion methodology paper and not repeated here; this paper focuses on
the per-revision trajectory.

\section{Setup: a 30-Revision Developmental Panel}
\label{sec:setup}

We run the spectral-and-screen recipe at 10 log-spaced revisions in each
of the three 1B-class models, for 30 mechanistic-interpretability runs in
total.

\paragraph{Revision schedules.} Pythia 1B's checkpoint set spans
\texttt{step1} ($\approx 2{\times}10^6$ tokens, essentially random
initialisation) through \texttt{step143000} ($\approx 300\mathrm{B}$
tokens, end of training). OLMo 1B's set spans
\texttt{step1000} (2B tokens) through
\texttt{step1454000} (3048B tokens). OLMoE 1B-7B's set spans
\texttt{step5000} (20B tokens) through \texttt{step1223842}
(5117B tokens). The
selected log-spaced revisions for each model are listed in
Table~\ref{tab:revisions}.

\begin{table}[t]
\centering
\caption{Revisions used for the 30-run developmental panel.
``Tokens'' is the cumulative training-token count published on the
respective HuggingFace model card; ``frac'' is tokens divided by the
model's final training-token count.}
\label{tab:revisions}
\small
\begin{tabular}{l r r r}
\toprule
Model & Revision label & Tokens (approx.) & frac of training \\
\midrule
Pythia 1B           & \texttt{step1}      & $2.1\,\mathrm{M}$    & $7\times10^{-6}$ \\
                    & \texttt{step8}      & $8.4\,\mathrm{M}$    & $3\times10^{-5}$ \\
                    & \texttt{step64}     & $6.7\times10^{7}$    & $2\times10^{-4}$ \\
                    & \texttt{step256}    & $5.4\times10^{8}$    & $2\times10^{-3}$ \\
                    & \texttt{step512}    & $1.1\,\mathrm{B}$    & $4\times10^{-3}$ \\
                    & \texttt{step3000}   & $6.3\,\mathrm{B}$    & $2\times10^{-2}$ \\
                    & \texttt{step10000}  & $21\,\mathrm{B}$     & $7\times10^{-2}$ \\
                    & \texttt{step38000}  & $80\,\mathrm{B}$     & $2.7\times10^{-1}$ \\
                    & \texttt{step143000} & $300\,\mathrm{B}$    & $1.0$ \\
\midrule
OLMo 1B             & \texttt{step1000}    & $2\,\mathrm{B}$      & $7\times10^{-4}$ \\
                    & \texttt{step5000}    & $10\,\mathrm{B}$     & $3\times10^{-3}$ \\
                    & \texttt{step11000}   & $23\,\mathrm{B}$     & $8\times10^{-3}$ \\
                    & \texttt{step25000}   & $52\,\mathrm{B}$     & $1.7\times10^{-2}$ \\
                    & \texttt{step56000}   & $117\,\mathrm{B}$    & $3.8\times10^{-2}$ \\
                    & \texttt{step126000}  & $264\,\mathrm{B}$    & $8.7\times10^{-2}$ \\
                    & \texttt{step312000}  & $654\,\mathrm{B}$    & $2.1\times10^{-1}$ \\
                    & \texttt{step644000}  & $1350\,\mathrm{B}$   & $4.4\times10^{-1}$ \\
                    & \texttt{step1454000} & $3048\,\mathrm{B}$   & $1.0$ \\
\midrule
OLMoE 1B-7B         & \texttt{step5000}    & $20\,\mathrm{B}$     & $3.9\times10^{-3}$ \\
                    & \texttt{step10000}   & $41\,\mathrm{B}$     & $8.0\times10^{-3}$ \\
                    & \texttt{step25000}   & $104\,\mathrm{B}$    & $2.0\times10^{-2}$ \\
                    & \texttt{step100000}  & $419\,\mathrm{B}$    & $8.2\times10^{-2}$ \\
                    & \texttt{step200000}  & $838\,\mathrm{B}$    & $1.6\times10^{-1}$ \\
                    & \texttt{step400000}  & $1677\,\mathrm{B}$   & $3.3\times10^{-1}$ \\
                    & \texttt{step800000}  & $3355\,\mathrm{B}$   & $6.6\times10^{-1}$ \\
                    & \texttt{step1223842} & $5117\,\mathrm{B}$   & $1.0$ \\
\bottomrule
\end{tabular}
\end{table}

\paragraph{Precision.} Pythia 1B's early-checkpoint evaluation requires
\texttt{fp32}: baseline attention values at random-init scale fall below
fp16 representable range and the selectivity ratios degenerate to
undefined values. OLMo and OLMoE evaluations use \texttt{fp16}.

\paragraph{Checkpoint-availability caveat.} The earliest published
HuggingFace checkpoint differs by an order of magnitude across the three
models: Pythia 1B's \texttt{step1} is at $\approx 2$M tokens (training
fraction $7\times10^{-6}$), OLMo 1B's is at 2B tokens (training fraction
$7\times10^{-4}$), and OLMoE 1B-7B's is at 20B tokens (training fraction
$4\times10^{-3}$). The left margin of the per-revision figures in this
paper reflects this publication limit, not the absence of an underlying
trajectory; earlier training-fraction values for OLMo and OLMoE do not
exist on their respective HuggingFace repositories. We keep the x-axis
range uniform across the three panels in Figure~\ref{fig:dev3} so that
the cross-model visual comparison is fair.

\begin{figure}[t]
\centering
\includegraphics[width=0.85\linewidth]{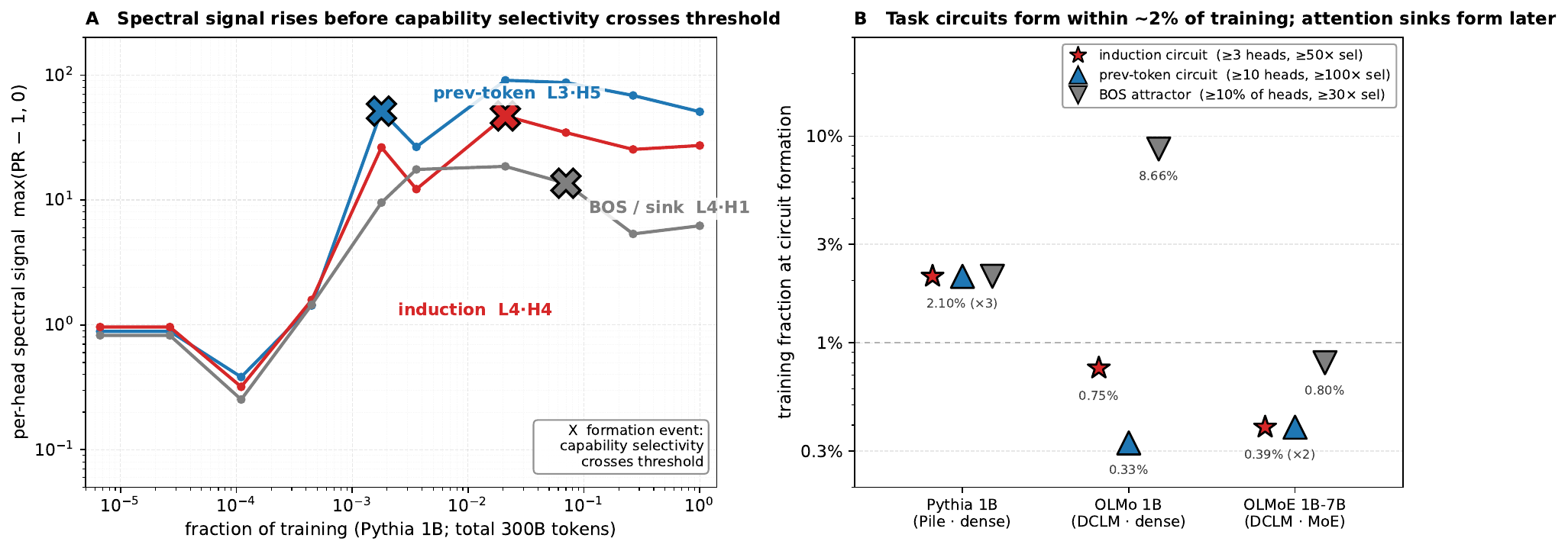}
\caption{\textbf{Capability circuits form early in pretraining and the per-head
spectral signal precedes formation.} Reproduced from the companion
methodology paper \citep{paper1_methodology}. \textbf{(A)} Per-head
spectral signal $\max(\prtxt_t-1,0)$ across training for three identified
heads in Pythia 1B (an induction head L4$\cdot$H4, a previous-token head
L3$\cdot$H5, a BOS-attractor head L4$\cdot$H1). X markers indicate the
formation event, the first revision at which the head's
capability-selectivity ratio exceeds its threshold. \textbf{(B)} Training
fraction at circuit formation across three 1B-class configurations:
capability circuits (induction, previous-token) form within the first
0.3--2.1\% of training in every configuration; the BOS attractor forms
later, with order-of-magnitude variation across configurations.}
\label{fig:headline}
\end{figure}

\paragraph{Reference figure.} Figure~\ref{fig:headline} (reproduced from
the methodology paper) provides the cross-configuration context: in every
panel and every model, capability circuits form within the first
$\sim$\,2\% of training, while the BOS attractor forms later and with
substantially more cross-configuration variation. The rest of this paper
unpacks that figure into the per-revision developmental detail that the
methodology paper summarises only in aggregate.

\section{Finding 1: L0--L1 Zero-BOS Is Universal Across Training}
\label{sec:f1}

The first developmental finding is also the simplest: in every one of the
three 1B-class models, at every one of the 10 sampled revisions, layers
0 and 1 produce \emph{zero} heads classified as
\textsc{first-token / bos} at the $\ge 30\times$ assignment threshold.
\citet{xiao2024streaming} observed this layer pattern on Llama-2-7B:
``the attention maps in the first two layers (layers 0 and 1) exhibit
the local pattern, with recent tokens receiving more attention; beyond
the bottom two layers, the model heavily attends to the initial tokens
across all layers and heads.'' Our developmental panel establishes that
this L0/L1 vs.\ L2+ separation is not just a property of final-checkpoint
Llama-2-7B but is present from the earliest sampled training revision of
every model we test, across two architecture families and two
pretraining datasets.

\paragraph{Per-layer counts: Pythia 1B.} Pythia 1B has 16 layers and 8
attention heads per layer (128 heads total). Per-layer counts of BOS-classified
heads at five representative revisions are shown in Table~\ref{tab:bos_pythia}.
The L0 and L1 columns hold at $0/8$ from \texttt{step1} (essentially
random initialisation) through \texttt{step143000} (300B tokens).
Other layers reach near-saturation: L4 transitions from $0/8$ at step3000
to $7/8$ at step10000 to $8/8$ at the final checkpoint.

\begin{table}[t]
\centering
\caption{Per-layer BOS-classified head counts in Pythia 1B at five
representative revisions. L0 and L1 hold at zero from random initialisation
through end of training, while L4 and later layers saturate.}
\label{tab:bos_pythia}
\small
\begin{tabular}{l r r r r r l l l}
\toprule
revision & L0 & L1 & L2 & L3 & L4 & L5--L8 & L9--L12 & L13--L15 \\
\midrule
\texttt{step1}      & 0/8 & 0/8 & 0/8 & 0/8 & 0/8 & 0,0,0,0 & 0,0,0,0 & 0,0,0 \\
\texttt{step512}    & 0/8 & 0/8 & 0/8 & 0/8 & 0/8 & 0,0,0,0 & 0,0,0,0 & 0,0,0 \\
\texttt{step3000}   & 0/8 & 0/8 & 0/8 & 0/8 & 0/8 & 0,1,2,1 & 0,4,1,3 & 1,1,0 \\
\texttt{step10000}  & 0/8 & 0/8 & 0/8 & 0/8 & 7/8 & 3,3,3,2 & 2,4,4,3 & 3,2,2 \\
\texttt{step143000} & \textbf{0/8} & \textbf{0/8} & 0/8 & 0/8 & 8/8 & 7,7,6,7 & 7,7,7,5 & 4,4,5 \\
\bottomrule
\end{tabular}
\end{table}

\paragraph{Per-layer counts: OLMo 1B.} OLMo 1B has 16 layers and 16 attention
heads per layer (256 heads total). The L0/L1 zero-BOS floor holds across all
sampled revisions (Table~\ref{tab:bos_olmo}). L2 transitions sharply between
revisions \texttt{step56000-tokens117B} (0/16 in L2) and
\texttt{step126000-tokens264B} (11/16 in L2); we revisit this DCLM-dense
phase-transition signature in Section~\ref{sec:f2}.

\begin{table}[t]
\centering
\caption{Per-layer BOS-classified head counts in OLMo 1B at five
representative revisions. L0 and L1 hold at zero; the L2--L4 transition
between \texttt{tokens117B} and \texttt{tokens264B} reflects the
DCLM-dense sharp transition documented in Section~\ref{sec:f2}.}
\label{tab:bos_olmo}
\small
\begin{tabular}{l r r r r r l l l}
\toprule
revision (label-tokens) & L0 & L1 & L2 & L3 & L4 & L5--L8 & L9--L12 & L13--L15 \\
\midrule
\texttt{step1000-2B}     & 0/16 & 0/16 & 0/16 & 0/16 & 0/16 & 0,0,0,0 & 0,0,0,0 & 0,0,0 \\
\texttt{step25000-52B}   & 0/16 & 0/16 & 0/16 & 0/16 & 0/16 & 0,0,0,0 & 0,0,0,0 & 0,0,0 \\
\texttt{step56000-117B}  & 0/16 & 0/16 & 0/16 & 3/16 & 9/16 & 7,0,0,0 & 0,0,0,0 & 0,0,0 \\
\texttt{step126000-264B} & 0/16 & 0/16 & 11/16 & 11/16 & 15/16 & 13,15,15,13 & 15,16,16,12 & 13,9,6 \\
\texttt{step1454000-3048B} & \textbf{0/16} & \textbf{0/16} & 14/16 & 16/16 & 16/16 & 15,16,16,15 & 16,16,15,15 & 16,14,7 \\
\bottomrule
\end{tabular}
\end{table}

\paragraph{OLMoE 1B-7B.} OLMoE 1B-7B shows the same pattern at all 10
revisions: zero BOS-classified heads at L0 or L1 at every revision. We
omit the full per-layer table for space; it tracks
Table~\ref{tab:bos_olmo} qualitatively.

\paragraph{Interpretation.} L0 and L1 hold at zero from random
initialisation through trillions of tokens; the model never crosses that
floor at any point during training. This is an \emph{architectural floor},
not a learned outcome. A natural mechanistic reading is that L0's inputs
are token embeddings (not yet contextualised), so a head that attends to
position 0 would discard its content (the position-0 embedding is a fixed
function of one specific token), and L1's inputs are L0 outputs, which by
the L0 floor do not yet contain any cross-token integration. Whatever an
attention sink ``is,'' it is not something the model can or wants to do
before the residual stream has accumulated nontrivial cross-position
information. A complete mechanistic theory of attention sinks must
explain why the gradient signal never installs BOS heads at L0/L1, at any
point during training, across pretraining corpora as different as Pile
and DCLM and across architectures as different as dense and MoE.

\subsection{Where sinks do emerge: mid-layer-outward, not bottom-up}
\label{sec:f1_midlayer}

Finding~1 establishes where sinks never form. The complementary question
is where they form \emph{first}. A natural prior is bottom-up: as the
residual stream accumulates cross-position information with depth, sinks
should appear at L2 (the first layer above the floor) and ascend. Tracking
per-layer BOS-classified head counts across the same 10 revisions shows
the opposite -- emergence is mid-layer-outward, and the layers just above
the floor are among the \emph{last} to acquire sinks, if ever:

\begin{itemize}
\item \textbf{Pythia 1B (Pile):} the first sinks appear at step3000
($\sim$2.1\% of training) in mid-to-late layers (L6--L8, L10--L14)
simultaneously. L0--L3 never acquire a sink at any revision --- the
zero-BOS floor extends two layers deeper than the L0/L1 lower bound in
this model.
\item \textbf{OLMo 1B (DCLM):} the first sink appears at L4 (step5000,
10B tokens, $\sim$0.33\%). L2 --- one layer above the floor --- does not
acquire a sink until step126000 (264B tokens, $\sim$8.7\%), $\sim$26$\times$
later in tokens than L4.
\item \textbf{OLMoE 1B-7B (DCLM, MoE):} the first sinks appear at
L7--L10, L12--L13 (step5000, 20B tokens, $\sim$0.39\%). L2 does not acquire
a sink until step600000 (2516B tokens, $\sim$49\% of training), more than
$100\times$ later than the mid-layer sinks.
\end{itemize}

The naive ``sinks ascend from the bottom'' picture is wrong in every
model: sinks crystallise first in the middle of the network and propagate
outward, with the near-floor layers (L2, L3) trailing or never
participating. A mechanistic account of sink formation must explain not
only the L0/L1 floor but this mid-outward ordering --- the gradient
installs sinks where mid-depth content-mixing is already underway, not at
the shallowest layer that is formally eligible.

\paragraph{Capability heads share layers with sinks.} A final-checkpoint
cross-section confirms that capability circuits live inside the sink
territory rather than beside it: in every model, every layer that contains
an induction-selective head ($\geq 30\times$) also contains at least one
BOS-classified head, and there are no induction-only layers above the
floor. This is partly a base-rate consequence --- BOS-classified heads are
52--81\% of all heads --- but it sharpens the picture from Finding~1: the
capability circuit is not segregated into sink-free layers; it is
interleaved with the attention-sink population in the same mid-depth
layers where both emerge.

\paragraph{Cross-model timing co-varies with data and architecture.} The
absolute token count at which sinks first appear is later in Pile-trained
Pythia and earlier in the DCLM-trained models, but this difference
co-varies with both pretraining corpus and architecture and we do not
isolate the two; the claim we rest on is the within-model layer
\emph{ordering} (mid-outward), which is invariant across all three models
regardless of corpus or architecture.

\section{Finding 2: BOS-Attractor Formation Has Three Distinct Shapes}
\label{sec:f2}

The whole-model BOS-class fraction (fraction of all heads classified
\textsc{first-token / bos} at the $\ge 30\times$ assignment threshold)
follows three qualitatively distinct trajectories across the three models
(Table~\ref{tab:bos_shapes}).

\begin{table}[t]
\centering
\caption{Whole-model BOS-class fraction (\% of all heads classified
first-token at $\ge 30\times$ selectivity) by tokens trained.
Em-dashes indicate that no model checkpoint is sampled at that
token count for the respective model. The OLMo 1B jump from 7.4\% at 117B
to 70.3\% at 264B is discontinuous between two adjacent published
checkpoints.}
\label{tab:bos_shapes}
\small
\begin{tabular}{r r r r}
\toprule
Tokens & Pythia 1B & OLMo 1B & OLMoE 1B-7B \\
\midrule
$\sim$6B           & 10.9\% & ---           & ---     \\
20--23B            & ---    & 0.0\%         & 3.5\%   \\
$\sim$80B          & 46.1\% & ---           & ---     \\
104--117B          & ---    & 7.4\%         & 42.2\%  \\
264B               & ---    & \textbf{70.3\%} & ---   \\
300B / 419B        & 57.8\% & ---           & 50.8\%  \\
838B               & ---    & ---           & 69.9\%  \\
1350--1677B        & ---    & 80.9\%        & 71.5\%  \\
3048--3355B        & ---    & 80.9\%        & 74.2\%  \\
5117B              & ---    & ---           & 75.8\%  \\
\bottomrule
\end{tabular}
\end{table}

The three shapes:

\paragraph{Pythia 1B (Pile, dense): gradual monotonic ramp.} The BOS-class
fraction rises from $0\%$ at random init to $10.9\%$ at $\sim$6B tokens,
$46.1\%$ at $\sim$80B, and $57.8\%$ at the final checkpoint (300B). The
trajectory is monotonic and smoothly varying across log-spaced revisions
with no sharp transitions visible at the resolution of this checkpoint
grid.

\paragraph{OLMo 1B (DCLM, dense): sharp phase transition.} The BOS-class
fraction holds at $0\%$ through 52B tokens, rises to $7.4\%$ at 117B, and
\emph{jumps} to $70.3\%$ at 264B --- between two adjacent published
checkpoints. The fraction is then essentially saturated, reaching $80.9\%$
at 1350B and remaining there at 3048B. The jump between 117B and 264B is
the developmental signature of a sharp phase transition: a 63-percentage-point
increase between adjacent checkpoints on a $\log_2$-spaced grid.

\paragraph{OLMoE 1B-7B (DCLM, MoE): gradual monotonic ramp.} The BOS-class
fraction rises from $3.5\%$ at 20B to $42.2\%$ at 104B, $50.8\%$ at 419B,
$71.5\%$ at 1677B, and $75.8\%$ at the final 5117B checkpoint. The shape
is qualitatively like Pythia's (monotonic, smooth on the log-spaced grid)
but the final-checkpoint saturation level is higher.

\paragraph{Substantive claim.} OLMo 1B and OLMoE 1B-7B are trained on the
same DCLM data but produce \emph{opposite} BOS-attractor emergence shapes:
a sharp phase transition in dense OLMo and a gradual ramp in MoE OLMoE.
The architectural change (dense $\to$ MoE) both reduces the final
BOS-class fraction (by $\sim$10pp at the final available revision for
each, $80.9\%$ vs $75.8\%$) and \emph{smooths} the developmental dynamics.
The architectural effect on the trajectory shape is separate from the
architectural effect on the asymptotic magnitude --- the same intervention
changes both, in the same direction (less BOS), but along two distinct
axes (final magnitude and transition sharpness).

\paragraph{Implication.} A theory of attention-sink formation that
predicts \emph{when} sinks emerge cannot rely solely on data composition:
DCLM produces a sharp transition in dense OLMo but a gradual ramp in MoE
OLMoE. Architecture matters for the shape of the curve, not only for its
asymptote. This is one constraint a mechanistic account of sinks must
satisfy: the same data delivers two different formation curves under two
different architectures.

\section{Finding 3: Induction Emerges Before the BOS Attractor in DCLM Models}
\label{sec:f3}

Combining the induction-circuit identification at each revision with the
BOS-class fraction trajectory of the previous section, we can compare the
formation timing of the two phenomena directly. Table~\ref{tab:milestones}
records the first revision at which each of seven milestones is reached
in each model.

\begin{table}[t]
\centering
\caption{First-revision-crossing for each developmental milestone, in tokens.
Em-dashes indicate that the model does not reach that milestone at any
sampled revision.}
\label{tab:milestones}
\small
\begin{tabular}{l r r r}
\toprule
Milestone & Pythia 1B & OLMo 1B & OLMoE 1B-7B \\
\midrule
induction $\ge 3$ heads ($\ge 50\times$) & $\sim$6B    & 23B  & 20B \\
induction $\ge 5$ heads ($\ge 50\times$) & $\sim$6B    & ---  & 20B \\
BOS-class $\ge 10\%$ of heads            & $\sim$6B    & 264B & 41B \\
BOS-class $\ge 30\%$                     & $\sim$80B   & 264B & 104B \\
BOS-class $\ge 50\%$                     & 300B        & 264B & 419B \\
BOS-class $\ge 70\%$                     & ---         & 264B & 1677B \\
previous-token $\ge 30$ heads at $\ge 30\times$ & $\sim$6B & 117B & 20B \\
self $\ge 30$ heads at $\ge 30\times$    & $\sim$6B    & 52B  & 20B \\
\bottomrule
\end{tabular}
\end{table}

\paragraph{Induction reaches its final size early.} The induction circuit
reaches its final 3--6-head size within the first 6--25B tokens in all
three models, then stays flat or slightly contracts over the remaining
300B--5T tokens. The head \emph{count} stabilises this early; the head
\emph{identity} does not --- see Finding~6. By $\sim$6B tokens (Pythia), 23B
(OLMo), and 20B (OLMoE),
the screen already identifies $\ge\!3$ heads with induction selectivity
$\ge\!50\times$. Pythia and OLMoE reach the $\ge\!5$-head milestone at the
same revisions; OLMo does not reach 5 heads at the $\ge\!50\times$
threshold at any revision (its final circuit is 3 heads).

\paragraph{The BOS attractor saturates much later in DCLM.} In the two
DCLM models, the BOS attractor's 50\% milestone is reached an order of
magnitude later than the induction milestone:
\begin{itemize}
  \item \textbf{OLMo 1B}: induction at 23B, BOS-50\% at 264B ---
        \textbf{11.5$\times$ gap} in tokens.
  \item \textbf{OLMoE 1B-7B}: induction at 20B, BOS-50\% at 419B ---
        \textbf{21$\times$ gap}.
  \item \textbf{Pythia 1B}: induction and BOS-10\% co-emerge at $\sim$6B
        on this checkpoint grid; the BOS-50\% milestone is reached at
        300B, a 50$\times$ later token count, but the BOS-10\% co-emergence
        with induction at $\sim$6B means the gap between
        induction-emergence and \emph{onset} of BOS growth is below the
        resolution of this revision grid. Finer Pythia checkpoints
        between \texttt{step512} (1B) and \texttt{step3000} (6B) would
        be needed to resolve.
\end{itemize}

\paragraph{Headline claim.} Capability-circuit formation and
attention-sink formation are temporally separable in DCLM 1B models ---
not a single phase transition. The induction-formation event (Olsson's
phenomenon) and the BOS-attractor formation event (Xiao's phenomenon) are
two distinct training-time events, separated by 10--20$\times$ in tokens
in DCLM. This refines the ``induction phase transition'' framing: at 1B
scale on DCLM, the induction transition occurs well before the
attention-sink transition. Pythia 1B on Pile may have a smaller gap, but
its co-emergence on this grid is a checkpoint-resolution limit rather than
positive evidence for a single combined transition.

\paragraph{Previous-token and self-attention timing.} The previous-token
milestone (30 heads at $\ge\!30\times$) and the self-attention milestone
reach in three different orderings across the three models. In Pythia 1B,
all three milestones (induction, prev-token, self) co-emerge at $\sim$6B.
In OLMo, self appears at 52B and prev-token at 117B (both after induction
at 23B but well before BOS-50\% at 264B). In OLMoE, all three appear at
20B simultaneously. The pattern that survives across all three models:
the BOS-attractor saturation is later than the
induction/prev-token/self capability milestones, and the gap is largest
in dense DCLM.

\section[Finding 4: The Capability Screen Converges Within \texorpdfstring{$\sim$1\%}{1\%} of Training]{Finding 4: The Capability Screen Converges Within $\sim$1\% of Training}
\label{sec:f4}

If the developmental curves of the previous section say that the
induction circuit is essentially complete by 20B tokens, then the
capability-specific screen, applied at an intermediate checkpoint, should
recover most of the final-checkpoint circuit using only a small fraction of
the training budget. Table~\ref{tab:convergence} confirms this directly.

\begin{table}[t]
\centering
\caption{Recall of the final-checkpoint induction circuit by the
capability-specific screen at intermediate revisions. The
``fraction of total tokens for 67\% recall'' column shows the training
budget at which the screen has already recovered two-thirds of the
final-checkpoint induction circuit.}
\label{tab:convergence}
\small
\begin{tabular}{l r r r r}
\toprule
Model & 33\% recall & 67\% recall & 100\% recall & frac.\ for 67\% \\
      & (tokens)    & (tokens)    & (tokens)     & (of total) \\
\midrule
Pythia 1B (3-head)    & $0.5$B & $6$B  & $80$B   & $\sim$2.0\% \\
OLMo 1B (3-head)      & $2$B   & $10$B & $3048$B & $\sim$0.3\% \\
OLMoE 1B-7B (4-head)  & $20$B  & $41$B & $1677$B & $\sim$0.8\% \\
\bottomrule
\end{tabular}
\end{table}

\paragraph{Reading the table.} For OLMo 1B, the screen at the
\texttt{step5000} revision (10B tokens, training fraction $3\times10^{-3}$)
already identifies 2 of the 3 heads that comprise the final induction
circuit at the $\ge\!50\times$ threshold. For OLMoE 1B-7B, the same
2-of-3-or-2-of-4 milestone is reached at \texttt{step10000} (41B tokens,
0.8\% of total training). Pythia 1B is somewhat slower in fractional
terms (2\%), but its absolute token count for 67\% recall (6B) is
comparable to OLMo (10B) and OLMoE (41B); the fractional figure is
inflated mainly because Pythia 1B's total training budget is shorter
(300B vs 3T and 5T).

\paragraph{Convergence is to the circuit, not the model.} Two of the three
models reach the 100\% milestone (every final-checkpoint induction head
recovered by the screen) only at the final checkpoint or near it; OLMo
in particular only reaches 100\% recall at \texttt{step1454000}. The
100\% recall figure should not be confused with circuit \emph{completion}:
the screen continues to add and remove heads through training, and this
turnover is not marginal: strong early induction heads can decay out
entirely (OLMoE's two highest-selectivity heads at formation fall from
$>\!150\times$ to $\le\!3\times$), so the formation-time circuit overlaps the
final circuit by only $\sim$1/3 (Finding~6). The recall figures above count
only how quickly the \emph{final} heads appear, not the non-final heads
present early that later drop out. What converges at $\sim$1\%
of training is the \emph{core} of the circuit (67\% of its final size);
the last $1/3$ of the circuit is recruited more slowly, often through
threshold crossings of individual heads that hover near the $50\times$
boundary.

\paragraph{Practical consequence.} For circuit-discovery purposes, you
do not need the final model. An intermediate checkpoint at 10--40B tokens
already contains $\ge\!67\%$ of the final-checkpoint induction circuit.
This has implications for two settings: (i) early-stopping for
interpretability-driven training, where pretraining to $\sim$1\% of the
final budget is sufficient if circuit identification is the only goal;
(ii) interpretability budgets, where the bulk of identifiability is
achievable on intermediate checkpoints that already exist on
HuggingFace for the three studied models.

\subsection{The selectivity noise floor is stable across training}
\label{sec:f4_noisefloor}

The convergence result above concerns \emph{when} the screen finds the
circuit; a companion question concerns the \emph{threshold} the screen
uses. The methodology paper~\citep{paper1_methodology} introduces a
per-model null-selectivity calibration in which induction-selectivity is
computed against a random non-special target position rather than the true
induction target. The null distribution gives the per-model noise floor
for selectivity. For circuit-identification claims at any checkpoint,
the natural question is: does this noise floor drift over training, or
is a single calibration (at the final checkpoint) approximately
applicable throughout the developmental panel?

Direct measurement on Pythia~1B at five log-spaced revisions
(step1000, step5000, step20000, step80000, step143000) with 500
null draws per revision:

\begin{center}
\footnotesize
\begin{tabular}{lrrrr}
\toprule
Revision & null$_{p99}$ & null$_{\max}$ & real$_{\max}$ & $K@>$null$_{p99}$ \\
\midrule
step1000   & 1.47 & 7.33   & 125.47 & 20 \\
step5000   & 1.86 & 256.15 & 177.59 & 33 \\
step20000  & 2.08 & 74.16  & 192.35 & 33 \\
step80000  & 1.45 & 767.51 & 152.02 & 43 \\
step143000 & 1.98 & 482.27 & 181.60 & 33 \\
\bottomrule
\end{tabular}
\end{center}

\paragraph{null$_{p99}$ is stable across training.} The 99th-percentile
of the null distribution stays in the 1.45--2.08$\times$ band across
$\sim$140K training steps (a factor of 1.4 between extremes). A
threshold calibrated at the final checkpoint (null$_{p99} = 1.98$ in
Pythia~1B) applies, with no more than $\sim$0.5 units of variation, at
every intermediate revision tested. The methodology's threshold-based
circuit-membership rule does not require recalibration per checkpoint --
a single per-model threshold applies throughout training.

\paragraph{null$_{\max}$ is unstable across training.} The maximum null
selectivity wanders by two orders of magnitude (7.3 to 767) across the
same revision range. This confirms that null$_{\max}$ is driven by
isolated pathological null draws (concentrated heads whose preferred
target happens to align with a randomly-drawn null position) and is not
a robust statistic. Threshold rules based on null$_{p99}$ are
checkpoint-stable; rules based on null$_{\max}$ are not. The
methodology paper adopts null$_{p99}$ as the defended pre-filter
threshold; the per-checkpoint data here reinforces that choice.

\paragraph{The induction-active circuit grows over training.} The
count of heads with real induction-selectivity above the per-checkpoint
null$_{p99}$ grows from 20 at step1000 to 33--43 at step5000 and beyond.
Most circuit formation occurs in the first $\sim$5000 training steps
(Finding~3), and the threshold-based count plateaus at 33--43 heads
thereafter -- consistent with Finding~3's 0.3--2.1\% training fraction
at circuit formation. The threshold rule and the circuit-formation
timing are consistent measurements of the same underlying phenomenon,
computed by independent procedures.

\section{Finding 5: PR Rises at or Before Capability Selectivity}
\label{sec:f5}

The methodology paper \citep{paper1_methodology} states that the spectral
signal (per-head PR) is elevated \emph{at or before} the formation event
(the first revision at which the head's capability-selectivity ratio
exceeds threshold). This section unpacks that aggregate claim into the
per-revision detail, working through one trajectory in detail and then
listing the heads across all three models for which the pattern holds.

\paragraph{Worked example: Pythia 1B L4$\cdot$H4.} L4$\cdot$H4 is the
top-selective induction head in the final-checkpoint Pythia 1B model
(induction selectivity $181\times$). Its per-revision PR and induction
selectivity are given in Table~\ref{tab:l4h4}.

\begin{table}[t]
\centering
\caption{Per-revision PR and induction selectivity for the top induction
head L4$\cdot$H4 in Pythia 1B. PR rises sharply at 0.5B tokens; induction
selectivity does not cross the $50\times$ threshold until 6B tokens.
The bold rows mark the PR rise (top) and the induction-selectivity
threshold crossing (bottom).}
\label{tab:l4h4}
\small
\begin{tabular}{r r r}
\toprule
Tokens (B) & PR & Induction selectivity ($\times$) \\
\midrule
0.002 & 1.96  & 1.0   \\
0.008 & 1.96  & 1.0   \\
0.033 & 1.32  & 1.0   \\
0.134 & 2.58  & 1.0   \\
0.536 & \textbf{27.35} & 0.2   \\
1.073 & 13.21 & 3.9   \\
6.291 & 48.05 & \textbf{171.8} \\
20.971& 35.62 & 150.7 \\
79.691& 26.38 & 147.9 \\
299.892& 28.29 & 181.6 \\
\bottomrule
\end{tabular}
\end{table}

PR rises by an order of magnitude between $0.13$B and $0.54$B tokens
(from 2.58 to 27.35), and reaches its run-maximum at $6.3$B (PR = 48).
Induction selectivity is still below $4\times$ at $1.07$B tokens, and
does not cross the $50\times$ circuit threshold until $6.3$B --- a
$\sim 12\times$ lead of PR over induction selectivity in token count
($0.536$B vs $6.3$B).

\paragraph{Heads across all three models exhibiting the pattern.}
The same pattern (per-head PR elevated at or before the
capability-selectivity threshold crossing) holds for every
final-checkpoint induction head we sampled, across all three models:

\begin{itemize}
  \item \textbf{Pythia 1B}: L4$\cdot$H4, L7$\cdot$H0, L7$\cdot$H1.
  \item \textbf{OLMo 1B}: L2$\cdot$H11, L4$\cdot$H12, L12$\cdot$H8.
  \item \textbf{OLMoE 1B-7B}: L7$\cdot$H0, L9$\cdot$H8, L5$\cdot$H10,
        L12$\cdot$H14.
\end{itemize}

For each of these heads, the per-revision PR rises one or more checkpoints
before the per-revision induction selectivity crosses $50\times$. The
specific lead times in number of checkpoints vary
(1--2 checkpoints for induction; co-emergence at the same checkpoint for
previous-token; 2--4 checkpoints for BOS-attractor heads), but the
qualitative direction is consistent across all 10 heads and all three
models.

\begin{figure}[t]
\centering
\includegraphics[width=0.98\linewidth]{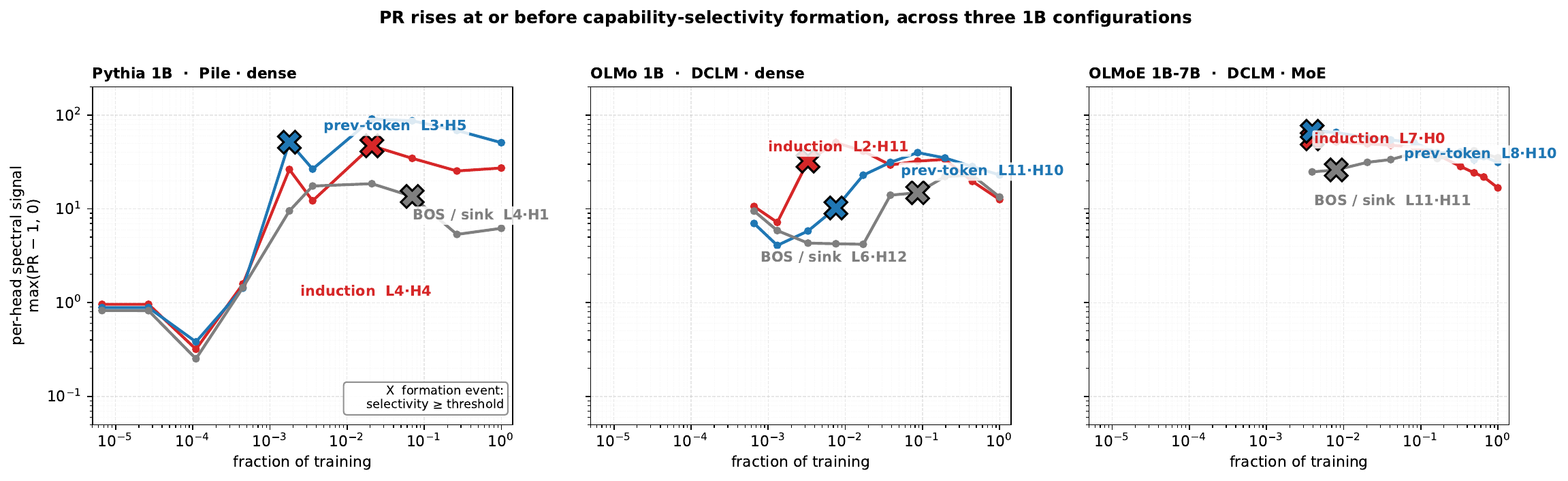}
\caption{\textbf{PR rises at or before capability-selectivity formation,
across three 1B-class configurations.} Per-head spectral signal at each
checkpoint, $\max(\prtxt_t-1,0)$ (the integrand of the PR-integral
ranking statistic, plotted per-checkpoint to preserve temporal structure;
same y-axis as Figure~\ref{fig:headline}A). For each of three 1B-class
configurations --- Pythia 1B (Pile, dense), OLMo 1B (DCLM, dense), and
OLMoE 1B-7B (DCLM, MoE) --- one top-selective head per capability class
is plotted: \textsc{induction} (red), \textsc{previous-token} (blue), and
\textsc{first-token / bos} (grey). Identifiers next to each curve give the
$(L,H)$ coordinates in the respective model. X markers indicate the
formation event --- the first checkpoint at which that head's
capability-selectivity ratio exceeds its threshold ($\ge\!50\times$ for
induction, $\ge\!100\times$ for previous-token, $\ge\!30\times$ for
first-token). In every panel and every curve the spectral signal is
elevated at or before the formation event. Lead times on this
10-checkpoint log-spaced grid: induction heads lead by 1--2 checkpoints
across all three models; previous-token heads co-emerge with their
capability selectivity (lead = 0); BOS-attractor heads lead by 2--4
checkpoints (longest in Pythia 1B and OLMoE, shortest in OLMo 1B where
the BOS-attractor transition is sharp; see Section~\ref{sec:f2}). The
cross-configuration consistency of the qualitative claim (PR elevated
at or before formation) is what the figure surfaces; the specific
lead-time numbers vary with the data, architecture, and the granularity
of the checkpoint grid. The empty space at the left of the OLMo and
OLMoE panels reflects the HuggingFace checkpoint-publication limit
discussed in Section~\ref{sec:setup}, not the absence of an underlying
trajectory; the x-axis range is kept uniform across panels for fair
visual comparison.}
\label{fig:dev3}
\end{figure}

\paragraph{Important scope statement.} The claim ``PR rises before
capability selectivity'' holds \emph{for the heads that end up
capability-selective}. The coupling is not specific to those survivors,
however: for the decay-out heads of Finding~6, per-head PR rises with their
induction selectivity and falls with it as they shed the pattern. It is
\emph{not} the same as the claim that
ranking heads by the PR integral $I(L,H)$ identifies capability-specific
heads. In the attention-sink-dominated 1B regime documented in
Sections~\ref{sec:f1}--\ref{sec:f2}, the top-$K$ heads by PR-integral
include L0/L1 generic content-dependent heads and BOS-class heads ahead
of the induction circuit, simply because the BOS-class heads have higher
\emph{sustained} PR through training. PR-integral is a general indicator
of specialized computation; the capability-specific screen of
Section~\ref{sec:method} disambiguates which specialization each head
is doing. We state this scope explicitly here rather than burying it
because it is the most common misreading of the headline figure: the
two-step pipeline (PR-integral $+$ class-specific screen) is necessary
for capability-specific identification in 1B+ models, even though the
PR signal alone is sufficient to flag the heads in
attention-sink-free smaller models.

\section{Finding 6: The Induction Circuit Turns Over During Training}
\label{sec:f6}

\begin{figure}[t]
\centering
\includegraphics[width=0.98\linewidth]{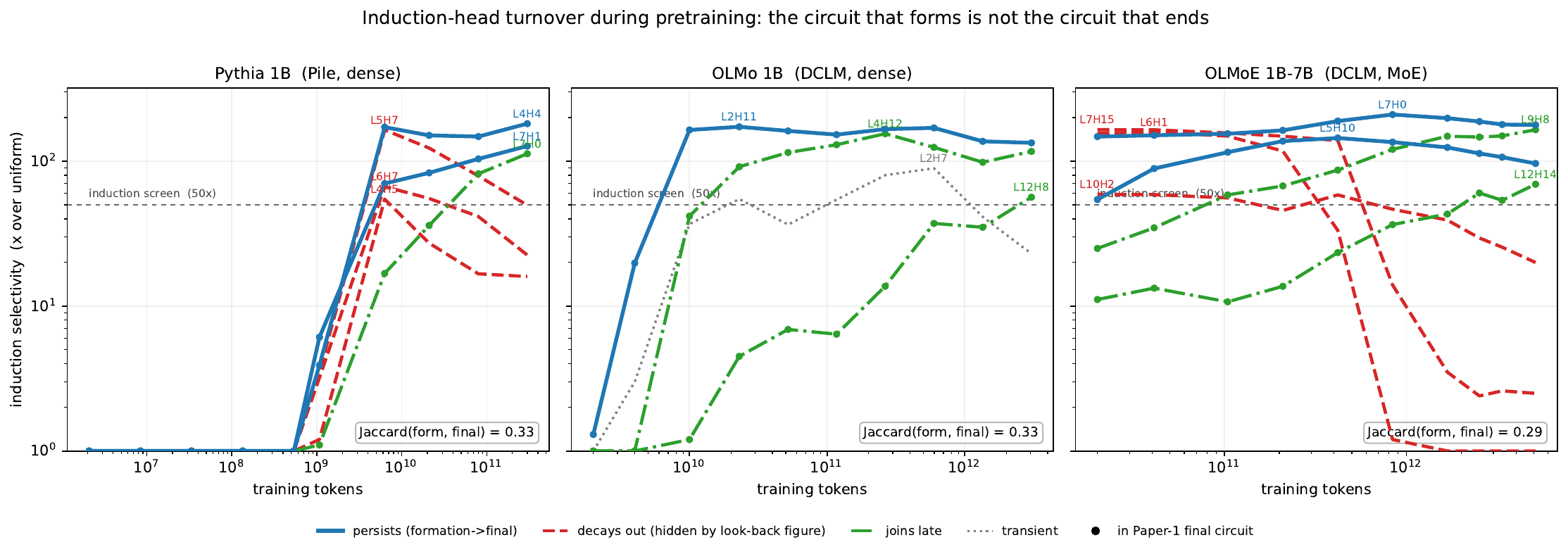}
\caption{\textbf{Induction-head turnover during pretraining.} Induction
selectivity (relative to a uniform-position baseline) of every head that passes
the $50\times$ screen at any revision, for each model. Heads that persist from
formation to the final checkpoint (blue), join late (green), or \emph{decay
out} after passing early (red); the red category is invisible to a figure that
plots only the final-circuit heads. Markers denote membership in the
final-checkpoint causal circuit \citep{paper1_methodology}. The
formation-revision circuit overlaps the final circuit by only $\sim$1/3
(Jaccard, inset in each panel).}
\label{fig:turnover}
\end{figure}

Finding~4 measured how quickly the \emph{final} induction heads appear --- a
recall measure. It is silent on the complementary question: at an intermediate
revision, are the heads passing the induction screen \emph{only} the eventual
final heads, or are others present that do not survive? Screening every head
independently at each revision --- rather than tracking the final-checkpoint
circuit backward --- answers this, and the answer is that the induction
circuit's membership turns over substantially during training
(Figure~\ref{fig:turnover}).

In all three models the induction \emph{function} appears early (Pythia~1B at
$\sim$6B tokens, OLMo~1B at $\sim$10B, OLMoE at or before its first published
checkpoint, 20B), but the heads implementing it change over the remainder of
training. Taking the \emph{formation circuit} to be the heads passing the
screen at the first revision any head does, its Jaccard overlap with the final
circuit is $0.33$ (Pythia), $0.33$ (OLMo), and $0.29$ (OLMoE): formation and
final circuits share only about a third of their combined heads, and this overlap stays in
$0.2$--$0.5$ across screen thresholds from $30\times$ to $100\times$ (OLMo's
overlap reflects late \emph{accretion} --- heads join, none drop --- rather
than the head-swapping seen in Pythia and OLMoE). Turnover runs in both
directions. Heads are recruited late in all three models --- Pythia's
L7$\cdot$H0 is absent at formation and reaches $113\times$ at the final
checkpoint, OLMoE's L9$\cdot$H8 climbs $25\times\!\to\!165\times$, and OLMo's
L12$\cdot$H8 rises $1\times\!\to\!56\times$. And in the two models with
multi-head formation circuits, strong early heads decay out entirely: OLMoE's
two highest-selectivity heads at formation, L7$\cdot$H15 ($165\times$) and
L6$\cdot$H1 ($156\times$), fall to $3\times$ and $1\times$ by the end of
training, and Pythia's L5$\cdot$H7 --- co-leading at formation ($164\times$,
against L4$\cdot$H4's $172\times$) --- decays to the screen threshold. This is
not marginal threshold-hovering; it is dominant heads losing the induction
pattern.

\paragraph{Turnover is specific to functional circuits, not the sink.}
Running the same independent per-revision screen on the other attention
classes separates two developmental modes. The attention sink does \emph{not}
turn over: the first-token class only \emph{accretes}, growing from a handful
of heads at formation to dozens or hundreds ($7\!\to\!63$ in Pythia,
$11\!\to\!202$ in OLMo, $4\!\to\!176$ in OLMoE) with \emph{zero}
dropouts at any screen threshold in any model --- once a head becomes a sink it stays
one. The specialized circuits do the opposite: induction sheds heads (above),
and in the MoE the previous-token and self classes churn too --- self most
strongly, with 9--30 heads dropping across thresholds (for instance
$37\!\to\!44$, 22 dropping and 29 joining at $50\times$) --- whereas the two
dense models show only marginal churn in these classes (single-digit dropouts
at most). Turnover --- a head losing a learned pattern --- is
thus a property of specialized computation, not of the attention sink, and it
is broadest in the mixture-of-experts.

\paragraph{Two sides of one coin with Finding~4.} Finding~4's recall measure
is correct --- most final heads do appear within $\sim$1\% of training --- but
it measures only how fast the \emph{final} heads arrive. Finding~6 supplies the
precision side: at that same point the screened set is not yet the final set,
because it also contains strong heads that later drop out. A circuit-discovery
procedure run on an intermediate checkpoint therefore recovers most of the
final heads while over-including transient ones; the circuit's identity is
fixed only late in training.

\paragraph{Interpretation.} This refines the ``induction phase transition''
account \citep{olsson2022}, in which induction heads form at the
in-context-learning transition and persist. The transition installs the
\emph{function} early, but the specific heads carrying it keep turning over
well past it. The induction capability is developmentally stable; its
circuit-level implementation is not.

\subsection{Is the turnover causal? Head fate, not architecture}
\label{sec:f6-causal}

Selectivity turnover is a statement about which heads pass the screen, not
about which heads do causal work: a head can shed the induction \emph{pattern}
while still being load-bearing for induction. To test the causal reading we
group-ablate the turning-over heads --- at the formation revision and at the
final checkpoint --- with the per-head zeroing protocol of
\citet{paper1_methodology} (Section~\ref{sec:method}), against
same-layer matched-random controls (4--8 seeds), scoring synthetic-induction
top-1 accuracy. Table~\ref{tab:causal-turnover} reports the dissociation.

\begin{table}[t]
\centering
\caption{\textbf{Causal reading of the turnover.} Change in synthetic-induction
top-1 accuracy (percentage points) when a head group is zeroed, at the formation
revision and at the final checkpoint. $z$ is the final-checkpoint effect
standardized against same-layer matched-random controls (4--8 seeds): strongly
negative $z$ = causally load-bearing, $z\!\approx\!0$ or positive = indistinguishable
from a random same-layer ablation (inert). Pythia's decay-out heads go inert by
the end (clean causal turnover); OLMoE's stay causal --- but this tracks the
\emph{fate} of the decaying head (sink vs.\ duplicate-token), not the
architecture (see text).}
\label{tab:causal-turnover}
\small
\begin{tabular}{l l r r r}
\toprule
Model & Group (final selectivity fate) & form. & final & $z$ \\
\midrule
Pythia 1B   & decay-out $\to$ sink                                  & $-10.2$ & $+0.3$ (inert)        & $+1.0$ \\
(dense)     & final-circuit                                         & ---     & $-3.8$                & $-2.9$ \\
\midrule
OLMoE       & decay-out (mixed fate)                                & $-10.3$ & $-4.0$ (still causal) & $<0$   \\
1B-7B       & \quad L6$\cdot$H1 alone $\to$ dup ($156\times\!\to\!1\times$)  & ---     & $-3.1$                & $-5.5$ \\
(MoE)       & \quad L10$\cdot$H2 alone $\to$ sink ($59\times\!\to\!20\times$) & ---     & $-0.3$ (inert)        & $-1.1$ \\
            & late-joiners                                          & $-1.7$  & $-4.3$                & $<0$   \\
\midrule
OLMoE       & L3$\cdot$H15 (always dup, induction never $>\!1\times$) & ---   & $-3.2$                & $-5.6$ \\
\bottomrule
\end{tabular}
\end{table}

\paragraph{The split is set by head fate, not by architecture.} A head that
loses induction selectivity ends up in one of two destination classes, and its
causal persistence tracks that destination rather than the model it lives in:

\begin{itemize}
\item \textbf{Induction $\to$ attention sink (BOS).} The head starts dumping
      attention on the first token and does no induction-relevant work: it
      genuinely leaves the circuit and is \emph{causally retired}. Pythia's
      three decay-out heads all take this fate (\,$\to$ 83--184$\times$ BOS\,)
      and go inert as a group ($+0.3$pp at the final checkpoint, $z=+1.0$);
      OLMoE's one sink-fate decayer, L10$\cdot$H2 ($\to$585$\times$ BOS), is
      likewise inert ($-0.3$pp).
\item \textbf{Induction $\to$ duplicate-token.} The head moves to attending
      from the current token to its previous occurrence --- which is exactly
      the prior-occurrence-detection step of the induction composition
      \citep{olsson2022}. It has slid from the copy-output role to the
      dup-detection role \emph{within} the circuit, so it stays causal. OLMoE's
      two strongest decayers take this fate: L7$\cdot$H15 ($\to$150$\times$ dup)
      and L6$\cdot$H1 ($156\times\!\to\!1\times$ induction, $\to$226$\times$
      dup), the latter still dropping induction $-3.1$pp ($z=-5.5$) at the final
      checkpoint despite having collapsed out of the induction screen.
\end{itemize}

The decisive observation is \emph{within OLMoE alone}: its sink-fate decayer is
inert and its dup-fate decayers are causal. The ``dense turns over, MoE
accretes'' contrast is therefore an aggregate artifact of which fate each
model's heads happened to take --- Pythia's all became sinks, OLMoE's strongest
two became duplicate heads --- not a property of dense vs.\ mixture-of-experts.
We do not have a model that exhibits the induction$\to$dup fate other than
OLMoE, so we make no claim that the dup fate is MoE-specific (Limitations).

\paragraph{Duplicate-detection is causally inside the induction circuit.} The
cleanest evidence is a head that was \emph{never} an induction head.
L3$\cdot$H15 in OLMoE is a duplicate-token head for its entire trajectory
(duplicate selectivity $152\times\!\to\!181\times$; induction selectivity never
exceeds $1\times$). Ablating it drops synthetic-induction top-1 by $-3.2$pp
against a matched-random control of $-0.48\pm0.48$pp ($z=-5.6$). A head doing
only duplicate-detection, which the induction screen never flags, is causally
necessary for induction --- so a head that slides induction$\to$dup keeps its
causal role because it moved within the circuit, not out of it.

\paragraph{Consequence: the selectivity screen under-identifies the causal
circuit.} The $\ge\!50\times$ induction screen measures only the copy-output
($A\,B\ldots A\!\to\!B$) attention pattern. It is therefore blind to the
duplicate-detection step, and a head performing that step can drop out of the
screen while remaining load-bearing --- which is precisely why OLMoE's
selectivity-retired heads are still causal. The practical reading of Finding~6
is thus two-sided: at an intermediate checkpoint the screen \emph{over}-includes
transient heads that will decay to sinks, and at any checkpoint it
\emph{under}-includes heads doing the dup-detection half of the mechanism. The
selectivity-turnover phenomenon stands; its causal interpretation is set by
where each decaying head goes.

\section{Discussion}
\label{sec:discussion}

\subsection{The ``induction phase transition'' framing}

\citet{olsson2022} described the formation of induction heads as a sharp
phase transition co-located with a bump in the loss curve and the
emergence of in-context learning. The headline result of this paper is
that, at 1B scale and on DCLM data, the induction transition and the
attention-sink transition are \emph{separated by 10--20$\times$ in
tokens}. The induction circuit reaches its final size at 20--23B tokens
in both DCLM models; the BOS-attractor reaches its 50\% milestone at
264B (OLMo) and 419B (OLMoE). These are not the same training-time event,
even on the same data, and even at adjacent scales (1B active parameters
in both DCLM models).

Three implications follow. First, ``the induction phase transition'' as a
phrase is ambiguous: it can mean the induction-circuit formation event
(20B tokens on DCLM, $\sim$6B on Pile in our panel) or the BOS-attractor
phase transition (264B in OLMo, gradual in OLMoE and Pythia). These are
mechanically distinct events with different timings. Second, sharp
phase-transition character is a property of \emph{specific} (data,
architecture) configurations rather than a universal feature of attention
formation: dense $+$ DCLM produces a sharp BOS-attractor transition;
MoE $+$ DCLM produces a gradual ramp on the same data; dense $+$ Pile
produces a gradual ramp on different data. Third, the relevant
mechanistic claim is no longer ``a single transition near loss-curve
feature $X$'' but ``two distinct transitions whose token-resolved timing
and shape depend on data and architecture.''

\subsection{What a theory of attention sinks must explain}

A complete mechanistic account of attention sinks needs to satisfy at
least three observed constraints from our developmental panel:

\begin{enumerate}
\item \textbf{The L0/L1 zero-BOS floor} (Section~\ref{sec:f1}). No
      gradient signal installs BOS-attractor heads at L0 or L1 at any
      point during training, across all three models. This is a property
      of where in the network BOS can do useful work, not a property of
      the data that eventually produces sinks. A theory of sinks needs
      to predict that the first two layers are off-limits for sink
      installation.
\item \textbf{The data-dependent shape} (Section~\ref{sec:f2}, comparing
      Pythia and OLMo). At fixed architecture (dense) and scale (1B),
      DCLM produces a sharper transition and higher saturation than
      Pile. The data does not just change the asymptote; it changes the
      sharpness of the transition. A theory that predicts the asymptote
      from data alone (e.g., from BOS-token frequency in the corpus)
      is insufficient.
\item \textbf{The MoE smoothing-and-reduction effect}
      (Section~\ref{sec:f2}, comparing OLMo and OLMoE). At fixed data
      (DCLM) and scale (1B active parameters), the MoE architecture
      both reduces the asymptotic BOS-class fraction and smooths the
      dynamics from a sharp transition to a gradual ramp. The
      architectural intervention changes two things at once. A theory
      should ideally identify a single underlying mechanism that
      manifests as both reduction and smoothing.
\end{enumerate}

A natural candidate framework that captures (1) is the
``residual-stream null-space'' picture: at L0 the residual stream is one
token's embedding plus position; at L1 the residual stream is L0's
embedding plus the L0 attention output, which by (1) is itself
content-dependent and never sink-routed. Sink heads can only do useful
no-op work once the residual stream carries a token-mix from
$\ge\!2$ prior layers of attention. We do not develop this picture
quantitatively; it is the kind of theory the L0/L1 floor calls for.

\subsection{Bridge to parameter-space emergence}

This paper studies emergence in \emph{activation space}: PR is a
spectral statistic of per-head attention outputs over an evaluation
batch. The capability selectivity ratio is an averaged attention pattern
on the same batch. Both quantities are activation-side, not
weight-side.

A complementary program studies the same emergence event from the
parameter-side. \citet{xu2026spectraledge} tracks the singular-value
spectral-edge geometry of weight-matrix trajectories through training and
reports signal--noise geometry signatures at training-token counts that
coincide, in their panel, with capability-emergence events identified by
downstream evaluation. \citet{xu2026gradsens} relates this, at toy-model
scale, to the Linear Centroids Hypothesis \citep{walker2026lch}: per-task
gradient SVD surfaces 100--330$\times$ coupling between Linear Centroids
features and the spectral-edge dynamics, coupling that is otherwise
hidden by optimizer trajectories. The picture both papers develop is that
parameter-space and activation-space both register the emergence event,
and that the parameter-space signature can be read out without
constructing a task-specific evaluation.

\paragraph{An exploratory measurement.} As a modest first probe of the
bridge within our panel --- not a test of the SED program itself --- we
relate the activation-side per-head PR to a parameter-side per-head
spectral statistic: the singular spectrum of each head's output-projection
($W_O$) slice, summarised by its stable rank
($\lVert W\rVert_F^2/\lVert W\rVert_2^2$), the participation ratio of its
singular values, and its top singular value $\sigma_1$. At the final
checkpoint there is a modest positive association between activation PR
and weight stable rank (Spearman $\rho$ pooled over 640 heads $=+0.32$;
per model $+0.16$ Pythia, $+0.28$ OLMo, $+0.38$ OLMoE) --- heads whose
outputs are spectrally diverse in activation space tend to have
higher-rank, more spread-out output projections. The cleanest single
signal is in Pythia, where high-PR heads have a \emph{low} top singular
value ($\rho(\mathrm{PR},\sigma_1)=-0.55$): a head that writes through one
dominant direction produces low-diversity output, as expected.
Developmentally, the association is not an endpoint artifact --- heads
that gain more activation PR over training also gain more weight stable
rank (growth co-evolution $\rho=+0.30$ Pythia, $+0.39$ OLMo, $+0.14$
OLMoE), and the cross-sectional association strengthens or holds over
training. Two scope limits matter. First, the relationship is at the
\emph{general-computation} level, not the capability level: weight spectra
do not predict induction selectivity (all $|\rho|\le 0.29$), consistent
with PR being a general specialisation indicator rather than a
capability-specific one. Second, it is a growth-direction tendency, not a
tight per-head lockstep (per-head trajectory correlations are inconsistent
in sign across models), and it uses a per-head weight-structure proxy
rather than the full trajectory statistic of the SED program. Data:
\verb|cross_architecture/results/param_space_bridge*.json|.

If the two programs are studying the same event, several further questions
sharpen. (i) The L0/L1 floor in activation space presumably has a
parameter-space counterpart: L0 and L1 attention matrices should
\emph{not} develop the spectral signature SED associates with
sink-formation in later layers. (ii) The two distinct transitions
(induction vs BOS-attractor) we observe should map to two distinct
parameter-space transitions, with the BOS-attractor transition's sharp
DCLM-dense character visible in OLMo's weight spectrum. (iii) The
predictive 0.3--2\% training-fraction recall in Section~\ref{sec:f4}
should have a parameter-space analogue: SED's spectral statistic at the
same checkpoints should identify the same set of attention heads. Beyond the general-level
per-head association measured above, none of these sharper questions are
tested here; they are the empirical questions an integrated
activation-and-parameter-space theory would need to answer.

\subsection{Practical methodological consequences}

Three concrete recommendations follow from the developmental findings:

\begin{enumerate}
\item \textbf{Apply the capability-screen at intermediate checkpoints
      rather than only the final one.} The 67\% recall threshold is
      reached at 0.3--2\% of total training in all three models, so
      most of the circuit is identifiable from a checkpoint two orders
      of magnitude smaller in training-token count than the final
      checkpoint.
\item \textbf{Distinguish ``induction emergence'' from ``attention-sink
      emergence'' in any developmental analysis.} They are not the
      same event in DCLM 1B models. Conflating them confuses what is
      being studied.
\item \textbf{When BOS-class fraction exceeds $\sim$70\%, use the
      all-head capability-specific screen rather than best-class
      ranking.} Best-class ranking surfaces BOS heads ahead of
      capability heads in this regime, which is the wrong filter for
      capability-specific causal claims (see
      Section~\ref{sec:f5}'s scope statement).
\end{enumerate}

\section{Limitations}

\paragraph{Checkpoint resolution.} The 10 log-spaced revisions per model
are the published HuggingFace checkpoints; finer-grained checkpoints would
catch transitions our grid currently lumps together. Pythia 1B in
particular: the 6B-token co-emergence of induction with BOS-10\% may be
two distinct transitions at finer resolution. The OLMo 1B sharp transition
between 117B (BOS-7\%) and 264B (BOS-70\%) similarly is the
between-checkpoint gap on a $\log_2$-spaced grid; the underlying
transition could be sharper or smoother on a $\log_{10}$-spaced grid.

\paragraph{Seed.} Each of the three 1B models is trained once, with a
single random seed; we do not have within-architecture re-pretraining at
this scale (re-pretraining a 1B-class natural-text model is expensive in
compute). The findings should therefore be read as ``what these specific
runs produced,'' not as ``what the (model, data) configuration produces in
expectation.'' At smaller scale, six seeds of TS-51M
\citep{paper1_methodology} show within-architecture variability in which
heads implement a target capability; we cannot rule out a similar
trajectory-shape variability at 1B. This bears directly on Finding~6: the
turnover \emph{phenomenon} appears in all three independent runs and is robust
to the screen threshold, but the \emph{specific} heads that decay out or join
are expected to be run-dependent --- our claims there concern turnover rates
and the sink/circuit asymmetry, not the fate of any individual head.

\paragraph{Causal turnover.} The fate mechanism of
Section~\ref{sec:f6-causal} carries three caveats. First, only OLMoE exhibits
the induction$\to$duplicate-token fate; Pythia's decay-out heads all become
sinks and OLMo has no clean induction-head decay to classify, so we have $n=1$
model with the dup fate and cannot say whether it is MoE-, data-, scale-, or
idiosyncrasy-related. Consequently OLMo cannot serve as a dense-DCLM control
for the dup fate. Second, the claim is about the two fates (sink $\to$ retired,
duplicate $\to$ retained), not about the specific heads taking them, which are
run-dependent as above. Third, the synthetic-induction baseline is low at the
final checkpoint ($\sim$4--5\% top-1), where a single matched-random control
seed can swing several points; every final-checkpoint causal claim here is
therefore taken against a 4--8-seed control distribution rather than a single
draw.

\paragraph{Evaluation batch.} The capability-selectivity ratios are
computed on the synthetic induction batch (random tokens, structured
$A\,B\,\ldots\,A$ pattern). Whether the same trajectory timings hold on
natural-text induction-target positions is an open question. The
final-checkpoint cross-validation in \citet{paper1_methodology} indicates
the synthetic-batch-identified induction circuit is the circuit doing
induction on natural text in OLMoE; we do not repeat that validation
per-revision in this paper because it would require natural-text batches
on every checkpoint.

\paragraph{MoE forward-pass cost.} OLMoE 1B-7B's forward pass involves
top-8 expert routing across 64 experts, which makes per-revision
mech-interp roughly 2$\times$ more expensive than dense OLMo at the same
nominal parameter count. We sampled 8 OLMoE revisions rather than 10 for
this reason. Two missing revisions are below 20B tokens, which is below
the earliest published HuggingFace checkpoint anyway.

\paragraph{Capability classes.} We tracked induction, previous-token,
self, BOS-attractor, duplicate-token, and local. The class taxonomy is
not exhaustive; in particular, no class in this list captures composed
tasks like IOI. The cross-task generalisation of the capability-screen
recipe is the subject of the companion pattern-selectivity paper
\citep{paper3_circuits}.

\section{Conclusion}

Tracking induction, previous-token, and BOS-attractor head emergence at
10 log-spaced revisions in each of three 1B-class language models reveals
that what is sometimes called ``the induction phase transition'' is, on
DCLM data, two transitions: capability-circuit formation by 20--25B
tokens, and attention-sink formation 10--20$\times$ later. The L0/L1
zero-BOS floor holds across all 30 runs; BOS-attractor formation follows
three distinct shapes across the three (data, architecture) configurations;
the capability-specific screen converges to most of the final induction
circuit within $\sim$1\% of training; and the per-head PR signal is
elevated at or before capability-selectivity threshold crossings, for
every final-checkpoint induction head sampled. The picture that emerges
is of \emph{two distinct training-time transitions} whose timing and
shape depend on data and architecture in separable ways, with the L0/L1
zero-BOS floor as a stable architectural constraint underneath both.

\bibliography{refs}

\end{document}